\documentclass[10pt,journal,compsoc]{IEEEtran}

\usepackage{mathrsfs}
\usepackage{amsthm,amsmath,amssymb}
\usepackage{algorithm}
\usepackage{subfigure}
\usepackage{graphicx}
\usepackage[misc]{ifsym}
\usepackage{multirow}
\usepackage{url}
\usepackage{amsmath}
\usepackage{amsthm}
\usepackage{booktabs}
\usepackage{algorithm}
\usepackage{algorithmic}

\begin{document}

\title{Registration of multi-view point sets under the perspective of expectation-maximization}

\author{Jihua Zhu,~\IEEEmembership{Member,~IEEE,}
        Jing Zhang,
        Huimin Lu,
        and~Zhongyu Li
\IEEEcompsocitemizethanks{\IEEEcompsocthanksitem J. Zhu and Z. Li are with Lab of Vision Computing and Machine Learning, School of Software Engineering, Xi'an Jiaotong University, Xi'an 710049, P.R. China.\protect\\
E-mail: zhongyuli@xjtu.edu.cn
\IEEEcompsocthanksitem J. Zhang is with School of Statistics and Mathematics, Zhongnan University of Economics and Law, Wuhan 430073, P.R. China. \IEEEcompsocthanksitem H. Lu is with Kyushu Institute of Technology, Kitakyushu, Japan.}

\thanks{}}

\markboth{}%
{Shell \MakeLowercase{\textit{et al.}}: Bare Advanced Demo of IEEEtran.cls for IEEE Computer Society Journals}

\IEEEtitleabstractindextext{%
\begin{abstract}
Registration of multi-view point sets is a prerequisite for 3D model reconstruction. To solve this problem, most of previous approaches either partially explore available information or blindly utilize unnecessary information to align each point set, which may lead to the undesired results or introduce extra computation complexity. To this end, this paper consider the multi-view registration problem as a maximum likelihood estimation problem and proposes a novel multi-view registration approach under the perspective of Expectation-Maximization (EM). The basic idea of our approach is that different data points are generated by the same number of Gaussian mixture models (GMMs). For each data point in one point set, its nearest neighbors can be searched from other well-aligned point sets. Then, we can suppose this data point is generated by the special GMM, which is composed of each nearest neighbor adhered with one Gaussian distribution. Based on this assumption, it is reasonable to define the likelihood function including all rigid transformations, which requires to be estimated for multi-view registration. Subsequently, the EM algorithm is utilized to maximize the likelihood function so as to estimate all rigid transformations. Finally, the proposed approach is tested on several bench mark data sets and compared with some state-of-the-art algorithms. Experimental results illustrate its super performance on accuracy, robustness and efficiency for the registration of multi-view point sets.
\end{abstract}

}

\maketitle

\IEEEdisplaynontitleabstractindextext

%
\IEEEpeerreviewmaketitle

\section{Introduction}
\IEEEoverridecommandlockouts
\IEEEPARstart{P}{oint} set registration is a fundamental methodology in many domains, such as computer vision~\cite{yang2015go, lei2017fast}, robotics~\cite{jiang2019simultaneous, yu2015semantic}, and computer graphics ~\cite{aiger20084, dai2017bundlefusion}. The development of point scanning devices makes it possible to reconstruct the 3D object or scene models. Due to the limited view, most scanning devices can only scan a part of the object or scene from one viewpoint. For the 3D model reconstruction, multiple  point sets should be acquired from different viewpoints to cover the entire object or scene surface, and then unified into the common reference frame by the multi-view registration. Accordingly, the multi-view registration is a prerequisite for 3D model reconstruction. Given multiple point sets, the goal of multi-view registration is to estimate the optimal rigid transformation for each point set and transform them from a set-centered frame to the same coordinate frame.

The  point set registration problem has attracted immense attention, and many effective approaches have been proposed to solve this problem. Among these approaches, one of the most popular solutions is the iterative closest point (ICP) algorithm ~\cite{besl1992method, chen1992object}, which can achieve the pair-wise registration with good efficiency and accuracy. However, including the ICP algorithm, most of them can not directly solve the multi-view registration problem. Compared with the pair-wise registration problem, the multi-view registration is a more difficult problem and has comparatively attracted less attention. Although some approaches have been proposed to solve this difficult problem, most existing approaches are unable to appropriately explore available information for the accurate registration. For the multi-view registration, some approaches only establish point correspondences between one and some other point sets, which cannot fully explore available information for accurate registration. While, other approaches may blindly establish point correspondences between one and all other point sets, which may lead to amount of extra computation complexity.

To this end, this paper considers the alignment of multiple  point sets as a maximum likelihood estimation problem and proposes a novel multi-view registration approach under the perspective of EM. The basic idea of our approach is that different data points are generated from the same number of GMMs. More specifically, each point set sequentially represents the data points and other opposite point  sets are utilized to define many GMMs. To define the GMM for each data point, its nearest neighbors are searched from other opposite point sets and they are viewed as all centroids of the corresponding GMM to generate the data point itself. Therefore, it is reasonable to define the likelihood function, which includes all rigid transformations for the multi-view registration. To achieve the multi-view registration, the EM algorithm is therefore utilized to maximize the likelihood function so as to estimate all rigid transformations.

The remainder of this paper is organized as follows. Section II surveys related works on the registration of point sets. Section III formulates the multi-view registration problem under the perspective of EM. Following that is section IV, in which the proposed method is derived to solve the multi-view registration problem. In Section V, the proposed method is tested and evaluated on six bench mark data sets. In Section VI, the proposed method is applied to the scene reconstruction. Finally, some conclusions are presented in Section VII.

\section{Related work}
\indent This section only surveys existing works related to our proposed approach for multi-view registration. For convenience, we will use the terms point set and range scan interchangeably throughout this paper.

Due to the number of involved point sets, the registration problem can be divided into two sub-problems, the pair-wise registration and the multi-view registration. For the pair-wise registration, one of the most popular methods is the ICP algorithm, which can achieve pair-wise registration with good performance. But
it cannot deal with non-overlapping point sets. Besides, it belongs to the local convergent algorithms. To improve its performance, many ICP variants have been proposed for pair-wise registration~\cite{rusinkiewicz2001efficient}. For non-overlapping point sets, Chetverikov et al.~\cite{chetverikov2005robust} proposed the trimmed ICP algorithm, which introduces the overlap percentage to automatically trim non-overlapping regions for accurate registration. To address local convergence, the Genetic algorithm~\cite{lomonosov2006pre} or the particle filter~\cite{sandhu2009point} is integrated with the TrICP algorithm to search the desired results. For the efficiency, some point feature methods~\cite{lei2017fast,rusu2009fast} are proposed to provide good initial parameters for the TrICP algorithm or its variants. 

Recently, some GMM-based method approaches, such as CPD~\cite{myronenko2010point}, GMMReg~\cite{jian2010robust} and FilterReg~\cite{gao2019filterreg} were also proposed to solve the pair-wise registration. Both CPD and FilterReg represent one point set as GMM, then cast the pair-wise registration problem as a maximum likelihood estimation problem. While, GMMReg utilizes two Gaussian mixture models to represent both point sets and reformulate the pair-wise registration as the problem of aligning two Gaussian mixtures, where a statistical discrepancy measure between two GMMs is minimized by the EM algorithm. Although these approaches can achieve pair-wise registration with good accuracy and robustness, they are time-consuming due to the huge number of point correspondences required to be established. Besides, they are unable to directly solve the multi-view registration problem.

For the multi-view registration, the intuitive method is the alignment-and-integration method~\cite{chen1992object}, which sequentially aligns and integrates two point sets until all point sets are integrated into one model. This approach is simple but suffers from the error accumulation problem due to a large number of point sets. Then, Bergevin et al.~\cite{bergevin1996towards} proposed the first solution for multi-view registration. It organizes all point sets by a star-network and sequentially puts one point set in the center of star-network. For the center point set, it finds point correspondences from each other point sets and estimates the rigid transformation by the ICP algorithm. As this approach can only sequentially estimate the rigid transformation for one point set, it may be difficult to obtain the desired registration results. Besides, the ICP algorithm is unable to deal with non-overlapping regions, so this approach is difficult to obtain promising results. What's more, it becomes more and more easily to be trapped into a local minimum with the increase of point set number.

To account for non-overlapping regions, Zhu et al. ~\cite{zhu2014surface} proposed a coarse-to-fine approach for multi-view registration, which sequentially traverses and refines the rigid transformation of each point set. More specifically, this approach views the traversed point set and all other coarse aligned point sets as data set and model set, respectively. Then the trimmed ICP algorithm is utilized to refine the rigid transformation of each traversed point set. To avoid the local convergence, Tang et al.~\cite{tang2015hierarchical} proposed a hierarchical approach to multi-view registration under the perspective of the graph, where each node and edge denotes a single point set and a connection between two overlapped point sets with non-slap percentage, respectively. Based on the graph, hierarchical optimization is implemented on the edges, the loops, and the entire graph. As it requires to detect all small loops, this approach turns to be difficult or impossible if none loop exist.

 To address the sequential estimation, Krishnan et al.~\cite{krishnan2005global} proposed an optimization-on-a-manifold approach for the multi-view registration. This approach can simultaneously estimate all rigid transformation from the established correspondences between each pair of point sets. However, it is difficult to establish accurate point correspondences in practical applications. Accordingly, Mateo et al.~\cite{mateo2014bayesian} treat the pair-wise correspondences as missing data and proposed the approach for multi-view registration under the Bayesian perspective. Although this approach may be accurate, it requires to calculate huge number of latent variables. Before this approach is proposed, some other graph-based approaches~\cite{shih2008efficient,torsello2011multiview} were also proposed for the multi-view registration. The difference is that each edge denotes the pair-wise registration of two connected nodes. Then the graph optimization approach is performed to diffuse the registration error over a graph of adjacent point sets. Without the update of point correspondences, these approaches only transfer registration errors among graph nodes and are unable to really reduce the total registration errors.

Recently, Govindu and Pooja~\cite{govindu2013averaging} proposed the motion averaging algorithm to solve the multi-view registration problem. This algorithm can directly estimate multi-view registration results (global motions) from a set of pair-wise registration results (relative motions) by the motion averaging algorithm. Meanwhile, Arrigoni et al.~\cite{arrigoni2016global} introduced the low-rank and sparse (LRS) matrix decomposition to estimate global motions from a set of available relative motions. Compared with the motion averaging algorithm, the LRS method is more robust to some unreliable relative motions. However, to obtain the desired registration results, it requires more relative motions than that of motion averaging algorithm.
What's more, the reliability of each relative motions is different, but these two methods consider that their contributions to the motion averaging are equal, which inevitably decreases the registration performance. To address this issue, Guo et al.~\cite{guo2018weighted} proposed the weighted motion averaging algorithm to solve the multi-view registration problem. Meanwhile, Jin et al.~\cite{jin2018multi} proposed the weighted LRS algorithm for multi-view registration. As these two approaches can pay more attentions to reliable relative motions, they can achieve more accurate and robust registration results than their original methods.

More recently, Evangelidis et al.~\cite{evangelidis2017joint} proposed the JRMPC method, which assumes that all points are realizations of the same GMM and then casts the multi-view registration into a clustering problem. Therefore, an EM algorithm is derived to achieve the clustering and estimate all rigid transformations as well as GMM parameters. As this approach requires to estimate a huge number of parameters, it is time-consuming and easy to be trapped into a local minimum. To address this issue, Zhu et al.~\cite{zhu2019efficient} derived K-means algorithm to achieve the clustering and estimate rigid transformations for the multi-view registration. This approach requires to estimate less parameters, so it is more efficient and robust than JRMPC. However, the performance of these clustering-based approaches is seriously affected by the number of clusters, which is a presetting parameter. In practical applications, we are difficult to know its optimal value without enough prior information. Besides, the clustering of points leads to the information loss, which reduces the registration performance.

Although both our method and JRMPC utilize the EM algorithm to achieve multi-view registration, their principles are totally different. JRMPC assumes that all points are generated from a central GMM, which contains a huge number of components required to be estimated. Accordingly, it requires to estimate GMM's parameters as well as all rigid transformations for multi-view registration. While, our approach assumes that different data points are generated from the same number of GMMs, which utilize equal covariances and equal membership probabilities for all Gaussian components. Given a data point, one NN is existing in each other point set and it can be viewed as one centroid of the corresponding GMM to generate the data point itself. As all centroids of each GMM can be efficiently searched and assigned with equal covariance, our approach only requires to estimate all rigid transformations as well as one GMM's covariance.

\section{Problem formulation}
\begin{figure}[ht]
\centering
\includegraphics[scale=0.7]{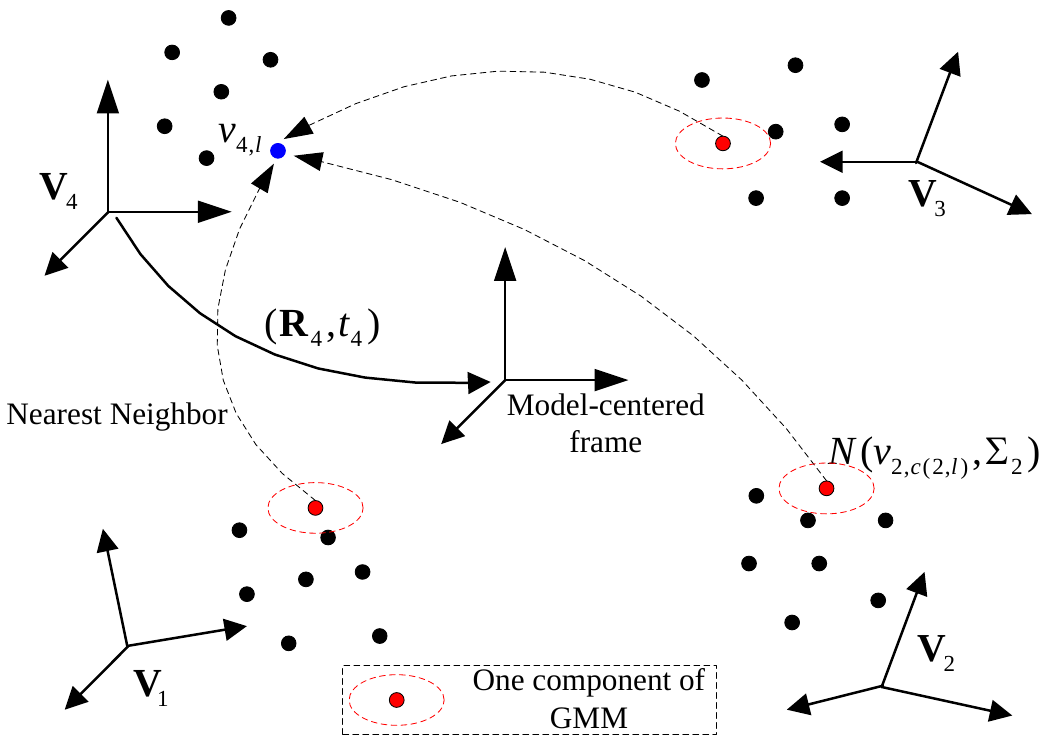}
\caption{The proposed method assumes that different well-aligned data points are generated from different GMMs, all of which are composed of $(M-1)$ equal components. For each data point in one point set, e.g. ${v_{4,l}} \in {{\mathbf{V}}}$, once rotated and translated from the set-centered coordinate frame to the model-centered frame, has one nearest neighbor in each other well-aligned point set. These nearest neighbors represent all centroids of the special GMM to generate the data point itself.}
\label{Fig:Schem}
\end{figure}

Let ${V} = \{ {{V}_i}\} _{i = 1}^M$ be the union of $M$ point sets and ${{V}_i} = [{v_{i,1}} \cdots {v_{i,l}} \cdots {v_{i,{N_i}}}] \in {\mathbb{R}^{3 \times {N_i}}}$ be $N_{i}$ points that belong to the $i$th point set. Given the model centered frame, the goal of multi-view registration is to estimate the rigid transformation ${\phi _i}:{\mathbb{R}^3} \to {\mathbb{R}^3}$ including a rotation matrix $\mathbf{R}_{i}$ and a translation vector $t_{i}$ for each point set, so as to transform them from a set-centered frame into the model centered frame. Fig. \ref{Fig:Schem} illustrates the principle of the proposed approach.

For one data point $v_{i,l}$ in the $i$th point set, there exist its nearest neighbors $\{ {v_{j,c(j,l)}}\} _{j = 1,j \ne i}^M$ in each other well-aligned point sets. Adhering with a Gaussian distribution, each nearest neighbor can be viewed as one centroid of a GMM, which contains $(M-1)$ components. Accordingly, we can assume the $i$th point set represents the data points and each of these data points is generated from one special GMM defined by its nearest neighbors in other opposite point sets. Besides, we use equal covariances and equal membership probabilities for all GMM components. Under this assumption, it is reasonable to formulate the joint probability of data point $v_{i,l}$ as follows:
\begin{equation}
P({v_{i,l}}) = \sum\limits_{j \ne i}^M {\frac{1}{{{M^{'}}}}} {\mathcal N} ({{\bf{R}}_i}{v_{i,l}} + {t_i};{{\bf{R}}_j}{v_{j,c(j,l)}} + {t_j},{\Sigma}),
\end{equation}
where $M^{'}=(M-1)$, ${{\mathbf{R}}_i} \in {\mathbb{R}^{3 \times 3}}$ and ${t_i} \in {\mathbb{R}^3}$ denote the rotation matrix and
the translation vector of the $i$th rigid transformation, respectively. For simplicity, we define the function
$\phi ({v_{i,l}}) = {{\bf{R}}_i}{v_{i,l}} + {t_i}$ for the rigid transformation $\{{{\bf{R}}_i},{t_i}\}$ imposed on the data point ${v_{i,l}}$.

To account for noise and outliers, it is essential to add an extra uniform distribution into the probability function:
\begin{equation}
P({v_{i,l}}) = wU(M) + (1 - w)\sum\limits_{j \ne i}^M {\frac{1}{{{M^{'}}}}} {\mathcal N} (\phi ({v_{i,l}});\phi ({v_{j,c(j,l)}}),{\Sigma}),
\label{eq:likeli}
\end{equation}
where $w$ is the parameter representing the ratio of outliers and ${\cal{U}}(M) = {1 \mathord{\left/ {\vphantom {1 M}} \right.\kern-\nulldelimiterspace} M}$ denotes the uniform distribution parameterized by the number of point sets involved in the multi-view registration.
As shown in Eq. (\ref{eq:likeli}), the probability function contains all rigid transformations for the multi-view registration. Accordingly, it requires to estimate these model parameters $\Theta = \{ \{ {{\mathbf{R}}_i},{{\mathbf{t}}_i}\} _{i = 1}^M, \Sigma\}$,
which can be achieved by maximizing the corresponding likelihood function.

\section{Multi-view registration approach under the perspective of EM}

As the estimation of model parameters can be achieved by maximizing the likelihood function, it is reasonable to utilize the EM algorithm. Therefore, it is necessary to define a set of hidden variables $Z = \{ {Z_{i,l}}|i \in [1,2,..,M],l \in [1,2,...,{N_i}]\}$, where ${Z_{i,l}} = c(j,l)$ means the observation $v_{i,l}$ is drawn from the Gaussian distribution ${\mathcal N}({v_{j,c(j,l)}},{\sum })$. Given all point sets $V$, model parameters can be estimated by maximizing the expected completed data log-likelihood function as follows:
\begin{equation}
\begin{array}{l}
\varepsilon (\Theta |V,Z) = {E_Z}[\log P(V,Z;\Theta )]\\
\quad \quad \quad \quad = \sum\limits_Z {P(Z|V,\Theta )} \log P(V,Z;\Theta )\\
\quad \quad \quad \quad = \sum\limits_Z {P(Z|V,\Theta )} \log P(V|Z,\Theta )P(Z;\Theta )
\end{array}
\end{equation}
As we utilize equal membership probabilities for all GMM components, $P(Z; \Theta )$ denotes the constant term. Therefore, $\varepsilon (\Theta |V,Z)$ can be reformulated as:
\begin{equation}
\varepsilon (\Theta |V,Z) = \sum\limits_Z {P(Z|V,\Theta )} \log P(V|Z,\Theta )
\label{eq:like1}
\end{equation}

For simplicity, it is reasonable to assume that all data points are independent and identically distributed. Accordingly, Eq. (\ref{eq:like1}) can be straightforwardly rewritten as:
\begin{equation}
\varepsilon (\Theta |V,Z) = \sum\limits_{i,l,j} {{\alpha _{i,l,j}}} \log P({v_{i,l}}|{Z_{i,l}} = c(j,l);\Theta ),
\end{equation}
where ${\alpha _{i,l,j}} = P({Z_{i,l}} = c(i,l)|{v_{i,l}},\Theta )$ denotes the posterior. By replacing the probability density of Gaussian distribution and ignoring constant terms, the objective function is reformulated as follows:
\begin{equation}
f(\Theta ) = - \sum\limits_{i,l,j} {{\alpha _{i,l,j}}} (\left\| {\phi ({v_{i,l}}) - \phi ({v_{j,c(j,l)}})} \right\|_\Sigma ^2 + \log \left| \Sigma \right|),
\label{eq:ffun}
\end{equation}
where $\left\| v \right\|_\Sigma ^2 = {v^T}{\Sigma ^{ - 1}}v$ and $\left| \Sigma \right|$ denotes the determinant of matrix $\Sigma$. For simplicity, we restrict each Gaussian distribution to the isotropic covariance, i.e., ${\Sigma} = {\sigma }^2{{\mathbf{I}}_3}$, where $\mathbf{I}_3$ denotes the $3 \times 3$ identity matrix. Therefore, Eq. (\ref{eq:ffun}) can be reformulated as:
\begin{equation}
f(\Theta) =  - \sum\limits_{i,l,j} {{\alpha _{i,l,j}}} (\frac{{\left\| {\phi ({v_{i,l}}) - \phi ({v_{j,c(j,l)}})} \right\|_2^2}}{{{\sigma ^2}}} + d\log {\sigma ^2})
\label{eq:obj4}
\end{equation}
where $d$ denotes point dimension, e.g. $d=3$ for range point.

 As these parameters $\left\{ {{{\mathbf{R}}_i}} \right\}_{i = 1}^M$ are of the Special Orthogonal $SO(3)$, particular care should pay attention to their estimation. Therefore, the multi-view registration can be formulated as the constrained optimization problem:
\begin{equation}
\left\{ {\begin{array}{*{20}{c}}
 {\mathop {\arg \max }\limits_\Theta f(\Theta )} \\
 {{\text{s}}{\text{.t}}{\text{.}}\quad {{\text{R}}_i}^T{{\text{R}}_i}{\text{ = I and}}\;\left| {{{\text{R}}_i}} \right| = 1,\;\forall i \in [1,...,M]}.
\end{array}} \right.
\end{equation}
This optimization problem can be solved by the EM algorithm, which is augmented with the establishment of point correspondences in E-step. Our approach can maximize the likelihood function to estimate all rigid transformations. We will refer to this approach as the \emph{expectation-maximization perspective for multi-view registration} (EMPMR), which achieves the multi-view registration by iterations. In each iteration, both E-step and M-step are included in the EMPMR.

\subsection{E-step}
In this step, we need to calculate the posterior probability of one data point $v_{i,l}$ generated from each component of the corresponding GMM. Before the calculation, it requires to specify the centroids of each GMM.

\subsubsection{E-Corresponding-step}
Given the parameter set $\Theta ^{k-1}$ obtained from the previous iteration, it is easy to transform all point sets into the same coordinate frame. Then, for one data point $v_{i,l}$ in the $i$th point set, it is required to find its corresponding point in each other opposite point set:
\begin{equation}
c(j,l) = \mathop {\arg \min }\limits_{h \in [1,2,..,{N_j}]} {\left\| {({{\bf{R}}_i}{v_{i,l}} + {t_i}) - \phi ({v_{j,h}})} \right\|_2},
\label{eq:nn}
\end{equation}
Eq. (\ref{eq:nn}) denotes the NN search problem, which can be efficiently solved by the $k$-d tree based method~\cite{nuchter2007cached}. For each data point in one point set, $M^{'}$ corresponding points are searched from other point sets and they are viewed as the centroids of the GMM to generate the data point itself.

\subsubsection{E-Probability-Step}
Given the centroid $v_{j,c(j,l)}$ and covariance $\Sigma$, it is easy to calculate the posterior probability of data point $v_{i,l}$ generated from the  Gaussian distribution ${\mathcal N}(v_{j,c(j,l)},\Sigma)$ as follows:
\begin{equation}
{\alpha _{i,l,j}} = \frac{{{\beta _{i,l,j}}}}{{\sum\nolimits_{j = 1,j \ne i}^M {{\beta _{i,l,j}} + \lambda } }},
\label{eq:pro}
\end{equation}
where $\lambda  = {\frac{{w{M^{'}}}}{{(1 - w)M}}}$  accounts for the outlier term and the notation ${\beta _{i,l,j}}$ denotes the probability density of Gaussian distribution
defined as:
\begin{equation}
{\beta _{i,l,j}} = \frac{1}{{{{(2\pi {\sigma ^2})}^{{\raise0.5ex\hbox{$\scriptstyle d$}
\kern-0.1em/\kern-0.15em
\lower0.25ex\hbox{$\scriptstyle 2$}}}}}}\exp ( - \frac{{\left\| {\phi ({v_{i,l}}) - \phi ({v_{j,c(j,l)}})} \right\|_2^2}}{{2{\sigma ^2}}}).
\end{equation}
Accordingly, ${a_{i,l,i}} = 1 - \sum\nolimits_{j = 1,j \ne i}^M {{a_{i,l,j}}} $ represents the posterior probability of the $v_{i,l}$ being an outlier.

\subsection{M-step}
Given current values of $\alpha_{i,l,j}$ and $c(j,l)$, this step requires to estimate all transformations $\{\mathbf{R}_{i},t_{i}\}_{i=1}^{M}$ by maximizing the function $f(\Theta)$. Although $M$ rigid transformations require to be estimated for multiple point sets, their estimation can be carried out independently for each point set. More specifically, we can alternative estimate one rigid transformation by setting other rigid transformations and the standard deviation $\sigma$ to their current values. Accordingly, the rigid transformation of
the $i$th point set can be estimated from the constrained problem:
\begin{equation}
\left\{ {\begin{array}{*{20}{c}}
 {\mathop {\arg \min }\limits_{{{\mathbf{R}}_i},{t_i}} (\sum\limits_{l = 1}^{{N_i}} {\sum\limits_{j = 1,j \ne i}^M {{\alpha _{i,l,j}}\left\| {({{\mathbf{R}}_i}{v_{i,l}} + {t_i}) - \phi ({v_{j,c(j,l)}})} \right\|_2^2} } )} \\
 {{\text{s}}{\text{.t}}{\text{.}}\quad {{\mathbf{R}}_i}^T{{\mathbf{R}}_i}{\text{ = I and}}\;\left| {{{\text{R}}_i}} \right| = 1,\;\forall i \in [1,...,M]} .
\end{array}} \right.
\label{eq:wICP}
\end{equation}
Eq. (\ref{eq:wICP}) denotes a weighted least square (LS) problem. As the parameter $\mathbf{R}_i$ is a special matrix, the Singular Value Decomposition (SVD) based method can be utilized to solve this weighted LS problem.

To facilitate analysis, the function $J(\mathbf{R}_{i},t_{i})$ is defined as:
\begin{equation}
J({{\mathbf{R}}_i},{t_i}) = (\sum\limits_{l = 1}^{{N_i}} {\sum\limits_{j = 1,j \ne i}^M {{\alpha _{i,l,j}}\left\| {({{\mathbf{R}}_i}{v_{i,l}} + {t_i}) - \phi ({v_{j,c(j,l)}})} \right\|_2^2} } ).
\label{eq:obj3}
\end{equation}
Taking the derivative of $J(\mathbf{R}_{i},t_{i})$ with respective to $t_{i}$, it is easy to obtain the following results:
\begin{equation}
\frac{{\partial J}}{{\partial {t_i}}} = \sum\nolimits_{l = 1}^{{N_i}} {\sum\nolimits_{j = 1,j \ne i}^M {{\alpha _{i,l,j}}[2{{\mathbf{R}}_i}{v_{i,l}} - 2\phi ({v_{j,c(j,l)}}) + 2{t_i}]} } .
\end{equation}
Let $\tfrac{{\partial J}}{{\partial {t_i}}} = 0$, the translation vector is estimated as:
\begin{equation}
{t_i} = \frac{{\sum\nolimits_{l = 1}^{{N_i}} {\sum\nolimits_{j = 1,j \ne i}^M {{\alpha _{i,l,j}}[{{\mathbf{R}}_i}{v_{i,l}} - \phi ({v_{j,c(j,l)}})]} } }}{{\sum\nolimits_{l = 1}^{{N_i}} {\sum\nolimits_{j = 1,j \ne i}^M {{\alpha _{i,l,j}}} } }}.
\label{eq:tvec}
\end{equation}
Then the $t_{i}$ in Eq. (\ref{eq:obj3}) can be replaced by Eq. (\ref{eq:tvec}) and the objective function is simplified as:
\begin{equation}
J({{\mathbf{R}}_i}) = \sum\nolimits_{l = 1}^{{N_i}} {\sum\nolimits_{j = 1,j \ne i}^M {{\alpha _{i,l,j}}\left\| {{{\mathbf{R}}_i}{p_{i,l}} - {q_{j,l}}} \right\|_2^2} },
\end{equation}
where
\begin{equation}
{p_{i,l}} = {v_{i,l}} - \frac{{\sum\nolimits_{l = 1}^{{N_i}} {\sum\nolimits_{j = 1,j \ne i}^M {{\alpha _{i,l,j}}{v_{i,l}}} } }}{{\sum\nolimits_{l = 1}^{{N_i}} {\sum\nolimits_{j = 1,j \ne i}^M {{\alpha _{i,l,j}}} } }},
\end{equation}
and
\begin{equation}
{q_{j,l}} = \phi ({v_{j,c(j,l)}}) - \frac{{\sum\nolimits_{l = 1}^{{N_i}} {\sum\nolimits_{j = 1,j \ne i}^M {{\alpha _{i,l,j}}\phi ({v_{j,c(j,l)}})} } }}{{\sum\nolimits_{l = 1}^{{N_i}} {\sum\nolimits_{j = 1,j \ne i}^M {{\alpha _{i,l,j}}} } }}.
\end{equation}
Accordingly, the rotation matrix ${{\mathbf{R}}_i}$ is estimated by minimizing the function $J({{\mathbf{R}}_i})$, which is expanded as follows:

\begin{equation}
\begin{gathered}
 {{\mathbf{R}}_i} = \mathop {\arg \min }\limits_{{{\mathbf{R}}_i}} \sum\limits_{l = 1}^{{N_i}} {\sum\limits_{j = 1,j \ne i}^M \begin{gathered}
 {\alpha _{i,l,j}}({p_{i,l}}^T{p_{i,l}} + {q_{j,l}}^T{q_{j,l}} \\
 - 2{p_{i,l}}^T{{\mathbf{R}}_i}{q_{j,l}}) \\
\end{gathered} } \hfill \\
 \quad \; = \mathop {\arg \max }\limits_{{{\mathbf{R}}_i}} \sum\limits_{l = 1}^{{N_i}} {\sum\limits_{j = 1,j \ne i}^M {{\alpha _{i,l,j}}({p_{i,l}}^T{{\mathbf{R}}_i}{q_{j,l}})} }. \hfill \\
\end{gathered}
\label{eq:objR}
\end{equation}

The optimization problem illustrated in Eq. (\ref{eq:objR}) has been well solved by the Singular Value Decomposition method ~\cite{nuchter2010study,myronenko2009closed}. Therefore, we only present the conclusion for the calculation of each rotation matrix ${{\mathbf{R}}_i}$ without proving.

$(1)$ Compute the matrix $\mathbf{H}$ and its singular value decomposition (SVD) results:
\begin{equation}
{\mathbf{H}} = \sum\limits_{l = 1}^{{N_i}} {\sum\limits_{j = 1,j \ne i}^M {\alpha _{i,l,j}}{{q_{j,l}}{p_{i,l}}^T} },
\end{equation}
\begin{equation}
\left[ {{\mathbf{U,\Lambda ,V}}} \right] = {\text{SVD}}({\mathbf{H}}).
\end{equation}

$(2)$ Estimate the rotation matrix:
\begin{equation}
{{\mathbf{R}}_i} = {\mathbf{V}}{{\mathbf{U}}^T},
\label{eq:rot}
\end{equation}
According to Eq. (\ref{eq:tvec}), it is easy to obtain the estimation of translation vector $t_i$. After the estimation of the $i$th rigid transformation, it requires to estimate the next rigid transformation until EMPMR obtains the desired results for multi-view registration.

Finally, when all rigid transformations have been updated, it requires to update the covariance matrix $\Sigma$ for the GMM. Take the derivative of $f(\Theta )$ with respective to $\sigma ^2$ and set it to $0$, $\Sigma$ can be updated as:
\begin{equation}
{\Sigma} = {\sigma }^2{{\mathbf{I}}_3},
\label{eq:sigma}
\end{equation}
where
\begin{equation}
{\sigma^2 } = \frac{{\sum\limits_{i,l,j} {{\alpha _{i,l,j}}} \left\| {({{\bf{R}}_i}{v_{i,l}} + {t_i}) - ({{\bf{R}}_j}{v_{j,c(j,l)}} + {t_j})} \right\|_2^2}}{{d\sum\limits_{i,l,j} {{\alpha _{i,l,j}}} }}.
\end{equation}
Obviously, the proposed method utilizes Eqs. (\ref{eq:nn}) and (\ref{eq:sigma}) to
specify the GMM to generate the data point $v_{i,l}$.

\subsection{Implementation}
Based on the above description, the proposed EMPMR approach is summarized in Algorithm 1. Similar to most registration approaches, EMPMR is a local convergent algorithm. For accurate registration, good initial guess of each rigid transformation should be provided in advance.

\begin{algorithm}[tb]
\caption{EMPMR approach}
\label{algorithm_SRLE}
\textbf{Input}: Point sets ${\mathbf{V}} = \{ {{\mathbf{V}}_i}\} _{i = 1}^M$, maximum iteration $K$,\\
\hspace*{0.38in}  initial guesses $\Theta^0$.\\
\textbf{Output}: $\Theta = \{ \{ {{\mathbf{R}}_i},{{\mathbf{t}}_i}\} _{i = 1}^M, \Sigma\}$.
\begin{algorithmic}[1] 
\STATE $k=0$;
\REPEAT
\STATE $k=k+1$;
\FOR{$(i = 1:M)$}
\STATE E-step
\STATE Build the correspondence $\{ {v_{i,l}},{v_{i,{c^k}(j,l)}}\}$ by Eq. (\ref{eq:nn});
\STATE Estimate the posterior probability ${\alpha^{k}_{i,l,j}}$ by Eq. (\ref{eq:pro});
\STATE M-step
\STATE Update ${\mathbf{R}}_i^k$ and $t_i^k$ by Eqs. (\ref{eq:tvec}) and (\ref{eq:rot}), respectively;
\STATE Update $\Sigma^{k}$ by Eq. (\ref{eq:sigma}).
\ENDFOR
\UNTIL{($\Theta $'s change is negligible)) or ($k>K$) }
\end{algorithmic}
\label{Algorithm1}
\end{algorithm}

\begin{table}[!t]
\centering
\renewcommand\arraystretch{1.5}
\caption{The total computation complexity of our approach}
\label{tab:1}
\begin{tabular}{|c|c|}
\hline \textbf{Operation} & \textbf{Complexity} \\
\hline $k$-d tree building & $O( \sum\nolimits_{i = 1}^M {{N_i}\log {N_j}})$\\
\hline E-Corresponding-step & $O(KM\sum\nolimits_{i = 1}^M {\sum\nolimits_{j = 1,j \ne i}^M {{N_i}\log {N_j}}})$\\
\hline E-Probability-Step & $O(KMM^{'}\sum\nolimits_{i = 1}^M {{N_i}})$ \\
\hline M-Step &
$O(KMM^{'}\sum\nolimits_{i = 1}^M {{N_i}} )$ \\ \hline
\end{tabular}
\label{Tab:Com}
\end{table}

\begin{table*}[!t]
\renewcommand\arraystretch{1.3}
\setlength{\tabcolsep}{7mm}
\centering
\caption{Details of data sets utilized in experiments}
\label{tab:2}
\begin{tabular}{|c|c|c|c|c|c|c|}
\hline 
{}& \textbf{Angel} & \textbf{Armadillo}& \textbf{Bunny} & \textbf{Buddha} & \textbf{Dragon} & \textbf{Hand}\\
\hline 
\textbf{Point sets} & 36 & 12 & 10 & 15 & 15 & 36\\
\hline 
\textbf{Points} &2347854 &307625 & 362272 &1099005 &469193 &1605575 \\ 
\hline
\textbf{Magnification} & ${10^1}$ & ${10^3}$ &${10^3}$ &${10^3}$ & ${10^3}$& ${10^1}$\\
\hline
\end{tabular}
\label{Tab:Data}
\end{table*}

\begin{table*}[!t]
\centering
\setlength{\tabcolsep}{4mm}
\renewcommand\arraystretch{1.2}
\caption{Comparison results on six data sets, where the numbers in bold denote the best performance.}
\label{tab:3}
\begin{tabular}{|c|c|c|c|c|c|c|c|c|c|c|}
\hline 
\multirow{2}{*}{}  & \multicolumn{2}{|c|}{\textbf{Initial}}&
\multicolumn{2}{|c|}{\textbf{K-means~\cite{zhu2014surface}}} &
\multicolumn{2}{|c|}{\textbf{MATrICP~\cite{govindu2013averaging}}} &
\multicolumn{2}{|c|}{\textbf{JRMPC~\cite{evangelidis2017joint}}} &
\multicolumn{2}{|c|}{\textbf{EMPMR}}\\
\hline 
&  $e_R$	& $e_t$ &  $e_R$	& $e_t$ &  $e_R$	& $e_t$ &  $e_R$	& $e_t$ &  $e_R$	& $e_t$\\
\hline \textbf{Angel} & 0.0312	& 2.0388 & 0.0079	& 0.9789 & 0.0146	& 2.8651 & 0.0092 &	1.9816 & \textbf{0.0033}	& \textbf{0.8809}\\
\hline \textbf{Armadillo} & 0.0331 & 2.5333 & 0.0094	& 2.3207 & 0.0336 &	4.2915 & 0.0171	& 3.0651 & \textbf{0.0058}	& \textbf{1.0908}\\
\hline \textbf{Bunny} & 0.0338 & 2.1260 & 0.0141 & 1.6088 & 0.0124	& 0.7181 & 0.0254 &	1.9551 & \textbf{0.0066}	& \textbf{0.3336}\\
\hline \textbf{Buddha} & 0.0371 & 1.6535 & 0.0208 & 1.0276 & \textbf{0.0120} & \textbf{0.8581} & 0.0226 &	0.9629 & 0.0123	& 1.0211\\
\hline \textbf{Dragon} & 0.0355	& 1.5216 & 0.0189	& 1.6140 & 0.0164	& 1.1037 & 0.0220 & 1.7466 & \textbf{0.0149} & \textbf{1.0160}\\
\hline \textbf{Hand} & 0.0823 &	0.4986 & 0.0067 & 0.6114 & 0.0371 & 1.3391 & 0.0103 & 0.7062 & \textbf{0.0064} & \textbf{0.3295}\\\hline
\end{tabular}
\label{Tab:Acc}
\end{table*}

\subsection{Complexity analysis}
This section analyzes the complexity of EMPMR. As EMPMR is proposed for the registration of multi-view point sets, the point set number $M$ and the total number of points $N_{i}$ or $N_j$ in each point set is the central quantity. For ease analysis, we suppose the iteration number of this approach is $K$. Before iteration, it requires to build $k-d$ tree to accelerate the NN search and the complexity is $O(N_{i}\log N_{i})$ for each point set. At each iteration, three operations are implemented to estimate one rigid transformation.

\textbf{E-Corresponding-step}. For each point $v_{i,l}$ in the $i$th point set, it is required to search its NN in each other point sets and the complexity is $O(\sum\nolimits_{j = 1,i \ne j}^M {\log {N_j}} )$. As there are $N_i$ in the $i$th point set and $M$ point sets involved in multi-view registration, the total complexity is $O(K\Sigma_{i=1}^M\Sigma_{j=1,j\neq i}^MN_{i}\log N_{j})$ for $K$ iterations.

\textbf{E-Probability-Step}. For each point $v_{i,l}$ in the $i$th point set, there are $M^{'}$ hidden variables. Accordingly, the proposed approach requires to calculate $M^{'}N_i$ hidden variables for the $i$th point set. Given $M$ point sets, the total complexity is $O(KM^{'}\Sigma_{i=1}^{M}N_{i})$ for $K$ iterations.

\textbf{M-Step}. Our method utilizes $M^{'}N_i$ point pairs and their corresponding hidden variables to estimate the $i$th rigid transformation. To estimate $M$ rigid transformations, the total complexity is $O(KM^{'}\Sigma_{i=1}^{M}N_{i})$ for $K$ iterations.

Therefore, Table \ref{Tab:Com} lists the total computation complexity for the estimation of $M$ rigid transformations.

\begin{figure}[!t]
\centering
\includegraphics[scale=0.6]{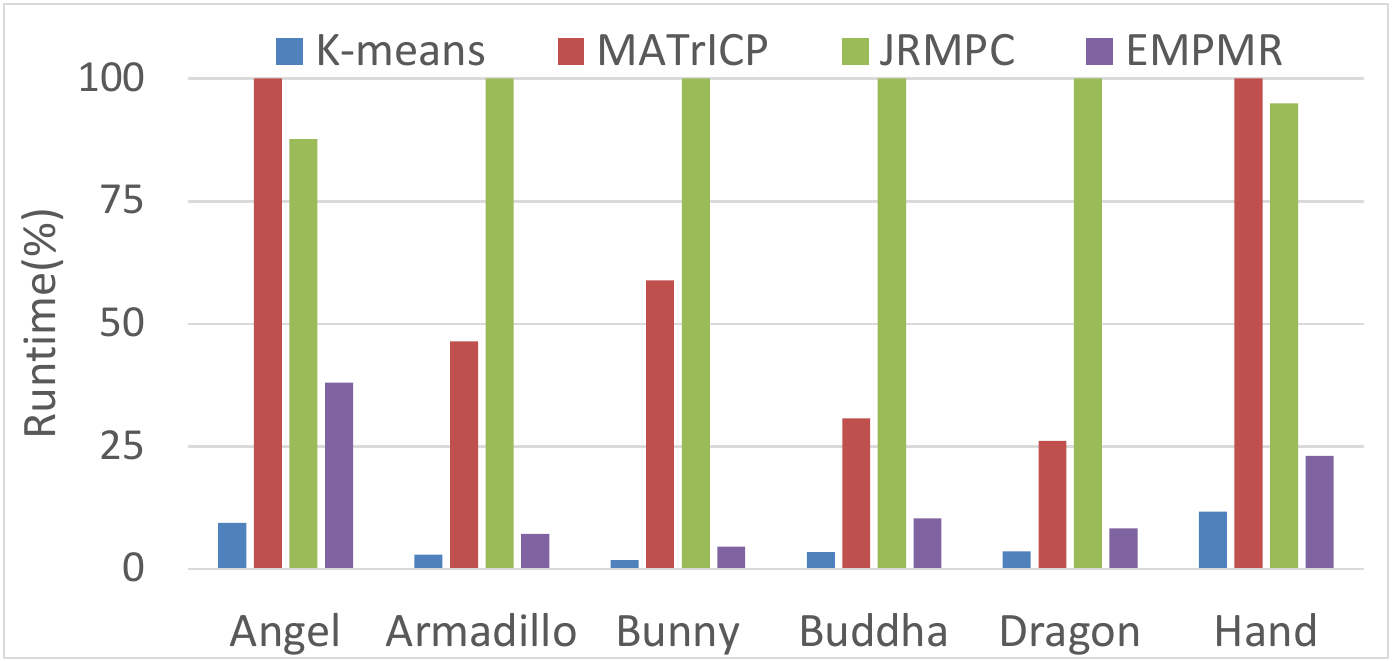}
\caption{Runtime comparison of different approaches tested on six data sets, where [10.3673, 3.2899, 2.6001, 3.8594, 4.026, 8.7427] minutes correspond to 100\% runtime of each data set.}
\label{Fig:Time}
\end{figure}

\section{Experiments}
In this section, EMPMR is tested and evaluated on six data sets, where four data sets are taken from the Stanford 3D Scanning Repository \cite{stanford3d} and the other two data sets were provided by Torsello ~\cite{torsello2011multiview}. Each of these data sets was acquired from one object model in different views and the multi-view point sets were provided along with the ground truth of rigid transformations for their registration. For accurate registration, all these data sets are magnified by different times. Table \ref{Tab:Data} illustrates some details of these data sets as well as the magnification times. To reduce the run time of registration, all data sets were uniformly down-sampled to around 2000 points per point set. In our method, the ratio of outliers is set to be $w= 0.01$.

To illustrate its performance, EMPMR is compared with three state-of-the-art approaches: the K-means based approach~\cite{zhu2019efficient}, the motion averaging approach with the TrICP algorithm~\cite{govindu2013averaging}, and the joint registration of multiple point clouds approach~\cite{evangelidis2017joint}, which are abbreviated K-means, MATrICP, and JRMPC, respectively. For different datasets, all approaches utilize the same setting for each parameter. Experimental results are reported in the form of the runtime, errors of rotation matrix and translation vector, where the error of rotation matrix and translation vector are defined as $e_{\mathbf{R}}=\frac{1}{M}\Sigma _{i=1}^M\| \mathbf{R}_{m,i}-\mathbf{R}_{g,i}\|_{F}$ and $e_{t}=\frac{1}{M}\Sigma _{i=1}^M\| t_{m,i}-t_{g,i}\|_{F}$, respectively. Here, $\{\mathbf{R}_{g,i},t_{g,i}\}$ and $\{\mathbf{R}_{m,i},t_{m,i}\}$ indicate the ground truth and the estimated one of the $i$th rigid transformation, respectively. All competed approaches utilize the $k$-d tree method to search the NN. Experiments are performed on a four-core 3.6 GHz computer with 8 GB of memory.

\begin{figure*}[!t]
\centering
\includegraphics[scale=1.16]{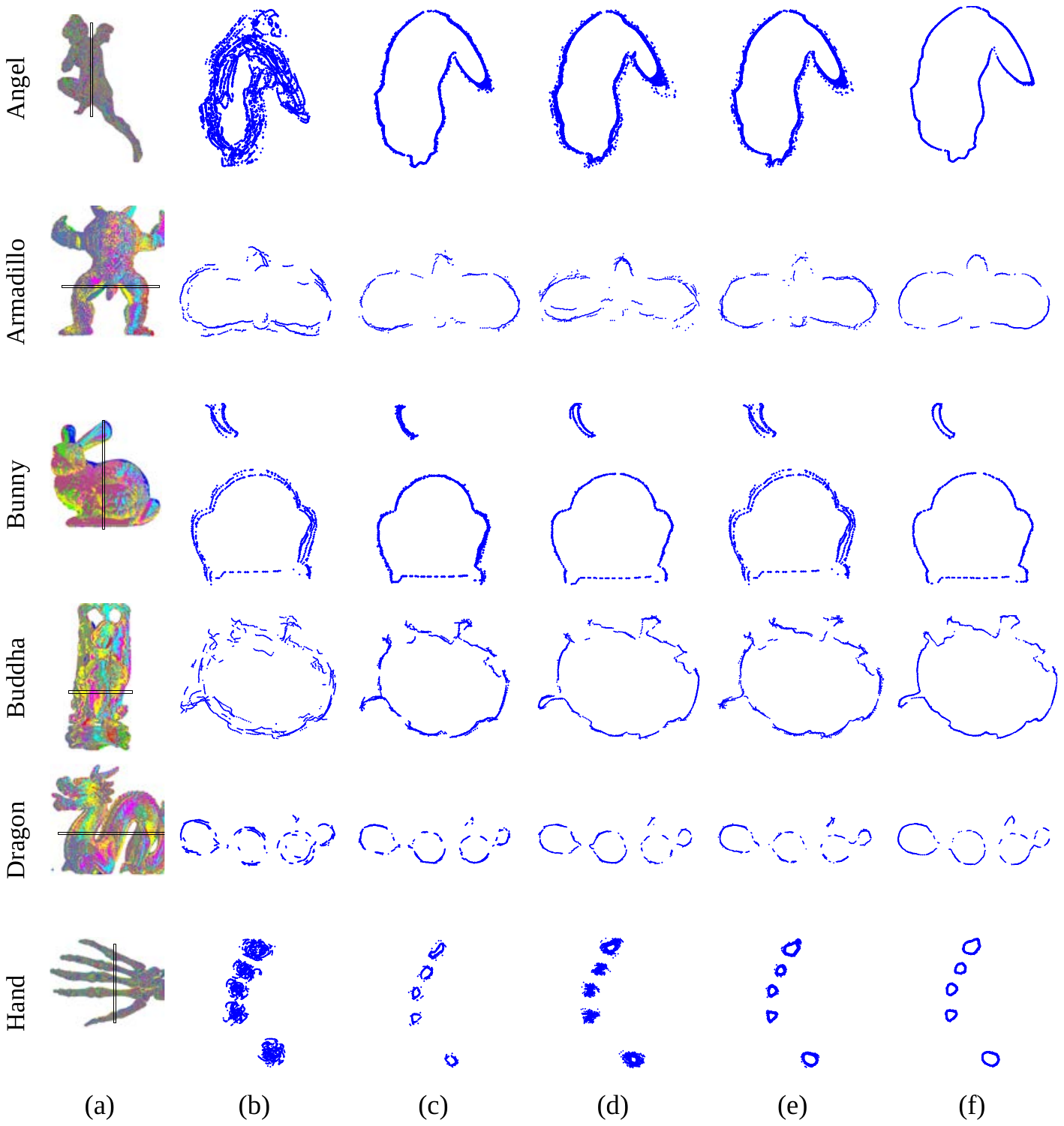}
\caption{Multi-view registration results in the form of cross-section. (a) Aligned 3D models. (b) Initial results. (c) Results of K-means method. (d) Results of MATrICP method. (e) Results of JRMPC method. (f) Our results.}
\label{Fig:Cros}
\end{figure*}

\subsection{Accuracy and efficiency}
Fig. \ref{Fig:Time} and Table \ref{Tab:Acc} illustrate multi-view registration results of different approaches tested on six data sets. To view these registration results in a more intuitive way, Fig. \ref{Fig:Cros} illustrates all multi-view registration results in the form of cross-section. As shown in Fig. \ref{Fig:Cros} and Table \ref{Tab:Acc}, EMPMR can always achieve the most accurate registration for all data sets, except for the Stanford Buddha. When refers to the efficiency, as displayed in Fig. \ref{Fig:Time}, K-means is the most efficient method and EMPMR is comparable with K-means, they are more efficient than the other two approaches. For the Stanford Buddha data sets, both MATrICP and EMPMR are able to obtain accurate registration results, where the former is a little better than the latter. 
For the remaining data sets, accurate registration results may be obtained by some other approaches. 
Although K-means is very efficient, it is less accurate than EMPMR, even less accurate than MATrICP for some data sets. As a method probabilistic approach, JRMPC can not obtain the accurate registration results as it is expected. 

\begin{table*}[!t]
\centering
\setlength{\tabcolsep}{2mm}
\renewcommand\arraystretch{1.2}
\caption{Comparison results (mean $\pm$ std.) on six data sets perturbed by the low Gaussian noise (SNR=50dB), where the numbers in bold denote the best performance. }
\begin{tabular}{|c|c|c|c|c|c|c|c|}
\hline 
\textbf{Method}                   &       & Angel              & Armadillo          & Bunny              & Buddha             & Dragon             & Hand               \\
\hline
\multirow{2}{*}{\textbf{K-means}} & $e_R$ & 0.0079$\pm$0.0004 & 0.0089$\pm$0.0006 & 0.0146$\pm$0.0005 & 0.0199$\pm$0.0007 & 0.0200$\pm$0.0008 & 0.0244$\pm$0.0105 \\
                                  & $e_t$ & 1.0750 $\pm$0.0636 & 2.3448 $\pm$0.1086 & 1.6936 $\pm$0.0628 & 1.0684 $\pm$0.0559 & 1.5158 $\pm$0.0724 & 0.7122 $\pm$0.0722 \\
\hline
\multirow{2}{*}{\textbf{MATrICP}} & $e_R$ & 0.0145 $\pm$0.0001 & 0.0304 $\pm$0.0066 & 0.0119 $\pm$0.0005 &\textbf{ 0.0122 $\pm$0.0003 }& 0.0160 $\pm$0.0004 & 0.0369 $\pm$0.0002 \\
                                  & $e_t$ & 2.8726 $\pm$0.0228 & 4.0146 $\pm$0.4889 & 0.6996 $\pm$0.0265 & \textbf{0.8424 $\pm$0.0168 }& 1.0534 $\pm$0.0449 & 1.3419 $\pm$0.0065 \\
\hline
\multirow{2}{*}{\textbf{JRMPC}}   & $e_R$ & 0.0090 $\pm$0.0002 & 0.0170 $\pm$0.0002 & 0.0223 $\pm$0.0002 & 0.0228 $\pm$0.0001 & 0.0218 $\pm$0.0005 & 0.0099 $\pm$0.0003 \\
                                  & $e_t$ & 1.9628 $\pm$0.0137 & 3.0534 $\pm$0.0162 & 1.7404 $\pm$0.0125 & 0.9694 $\pm$0.0042 & 1.7923 $\pm$0.0048 & 0.7059 $\pm$0.0022 \\
\hline
\multirow{2}{*}{\textbf{Ours}}    & $e_R$ &\textbf{ 0.0033 $\pm$0.0000 }&\textbf{ 0.0058 $\pm$0.0001 }&\textbf{ 0.0066 $\pm$0.0001 }& 0.0123 $\pm$0.0001 &\textbf{ 0.0149 $\pm$0.0001 }&\textbf{ 0.0064 $\pm$0.0000 }\\
                                  & $e_t$ &\textbf{ 0.8819 $\pm$0.0047 }&\textbf{ 1.0928 $\pm$0.0045 }&\textbf{ 0.3317 $\pm$0.0029 }& 1.0207 $\pm$0.0028 & \textbf{1.0159 $\pm$0.0027} & \textbf{0.3296 $\pm$0.0004}\\ 
\hline 
\end{tabular}
\label{Tab:50dB}
\end{table*}

\begin{table*}[!t]
\centering
\setlength{\tabcolsep}{2mm}
\renewcommand\arraystretch{1.2}
\caption{Comparison results (mean $\pm$ std.) on six data sets perturbed by the high Gaussian noise (SNR=25dB), where the numbers in bold denote the best performance.}
\begin{tabular}{|c|c|c|c|c|c|c|c|}
\hline 
\textbf{Method}                   &       & Angel              & Armadillo          & Bunny              & Buddha             & Dragon             & Hand               \\
\hline 
\multirow{2}{*}{\textbf{K-means}} & $e_R$ & 0.0080 $\pm$0.0005 & 0.0095 $\pm$0.0009 & 0.0150 $\pm$0.0007 & 0.0201 $\pm$0.0012 & 0.0213 $\pm$0.0011 & 0.0302 $\pm$0.0071 \\
                                  & $e_t$ & 1.0117 $\pm$0.0827 & 2.4698 $\pm$0.1109 & 1.7043 $\pm$0.0716 & 1.1212 $\pm$0.0866 & 1.3913 $\pm$0.1099 & 0.7447 $\pm$0.0523 \\
\hline 
\multirow{2}{*}{\textbf{MATrICP}} & $e_R$ & 0.0145 $\pm$0.0001 & 0.0289 $\pm$0.0073 & 0.0119 $\pm$0.0006 &\textbf{ 0.0121 $\pm$0.0004 }& 0.0162 $\pm$0.0005 & 0.0369 $\pm$0.0003 \\
                                  & $e_t$ & 2.8710 $\pm$0.0187 & 3.8764 $\pm$0.5655 & 0.7047 $\pm$0.0337 &\textbf{ 0.8433 $\pm$0.0188} & 1.0464 $\pm$0.0559 & 1.3417 $\pm$0.0063 \\
                                  \hline 
\multirow{2}{*}{\textbf{JRMPC}}   & $e_R$ & 0.0088 $\pm$0.0004 & 0.0168 $\pm$0.0006 & 0.0232 $\pm$0.0005 & 0.0229 $\pm$0.0003 & 0.0223 $\pm$0.0004 & 0.0103 $\pm$0.0007 \\
                                  & $e_t$ & 1.9069 $\pm$0.0526 & 3.0249 $\pm$0.0542 & 1.7402 $\pm$0.0320 & 0.9690 $\pm$0.0240 & 1.7817 $\pm$0.0272 & 0.7196 $\pm$0.0068 \\
                                  \hline 
\multirow{2}{*}{\textbf{Ours}}    & $e_R$ & \textbf{0.0033 $\pm$0.0003 }&\textbf{ 0.0059 $\pm$0.0005 }&\textbf{ 0.0070 $\pm$0.0005 }& 0.0128 $\pm$0.0009 &\textbf{ 0.0151 $\pm$0.0004} &\textbf{ 0.0072 $\pm$0.0003 }\\
                                  & $e_t$ &\textbf{ 0.8609 $\pm$0.0457 }& \textbf{1.0950 $\pm$0.0437 }&\textbf{ 0.3555 $\pm$0.0267 }& 1.0151 $\pm$0.0175 &\textbf{ 1.0092 $\pm$0.0244 }&\textbf{ 0.3551 $\pm$0.0051}\\
\hline 
\end{tabular}
\label{Tab:25dB}
\end{table*}

To achieve the multi-view registration, the K-means based method sequentially traverses each point set, then alternatively implements the operations of clustering and rigid transformation estimation. In the clustering operation, it uses the K-means algorithm to divide all aligned points into the preset number of clusters and utilizes one clustering centroid to represent all points in this cluster, which inevitably causes the loss of information. For the estimation of each rigid transformation, it aligns the point set to clustering centroids, where the correspondence of each data point is searched from all clustering centroids. Since the number of cluster centroids is far less than that of raw points in all point sets, this approach is very efficient but far from accuracy for the multi-view registration due to the information loss.

For the multi-view registration, MATrICP approach recovers all global motions from a set of relative motions estimated by the pair-wise registration. As the pair-wise registration problem is easier than the multi-view registration problem, the motion averaging algorithm can transform the latter into the former. For accurate registration, the motion averaging algorithm is expected to implement on as more relative motions as possible, which may introduce some unreliable relative motions. However, the motion averaging algorithm is sensitive to unreliable relative motions, and even one relative motion will lead to the failure of multi-view registration. As shown in Table \ref{Tab:Acc} and Fig. \ref{Fig:Cros}, for Angel, Stanford Armadillo and Hand datasets, MATrICP approach failures to obtain accurate registration results due to the input of unreliable relative motions. Besides, its efficiency is seriously reduced with the increase of point set number in multi-view registration. As illustrated in Fig. \ref{Fig:Time}, MATrICP is more efficient than JRMPC for multi-view registration of these data sets with small number of point set. Then it turns to be less efficient than JRMPC, when it is tested on other two data sets including a large number of point sets.

\begin{figure*}[t]
\centering
\subfigure[]{
\includegraphics[scale=0.53]{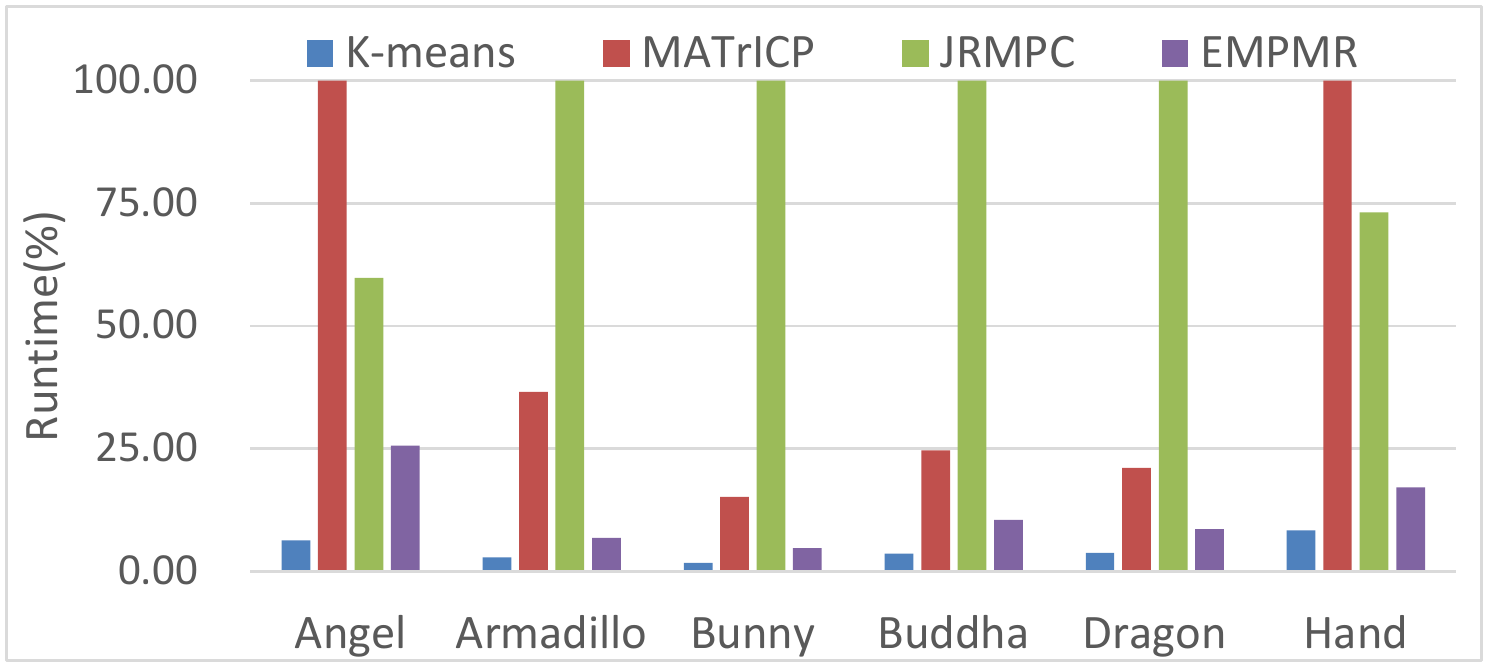}
}
\subfigure[]{
\includegraphics[scale=0.53]{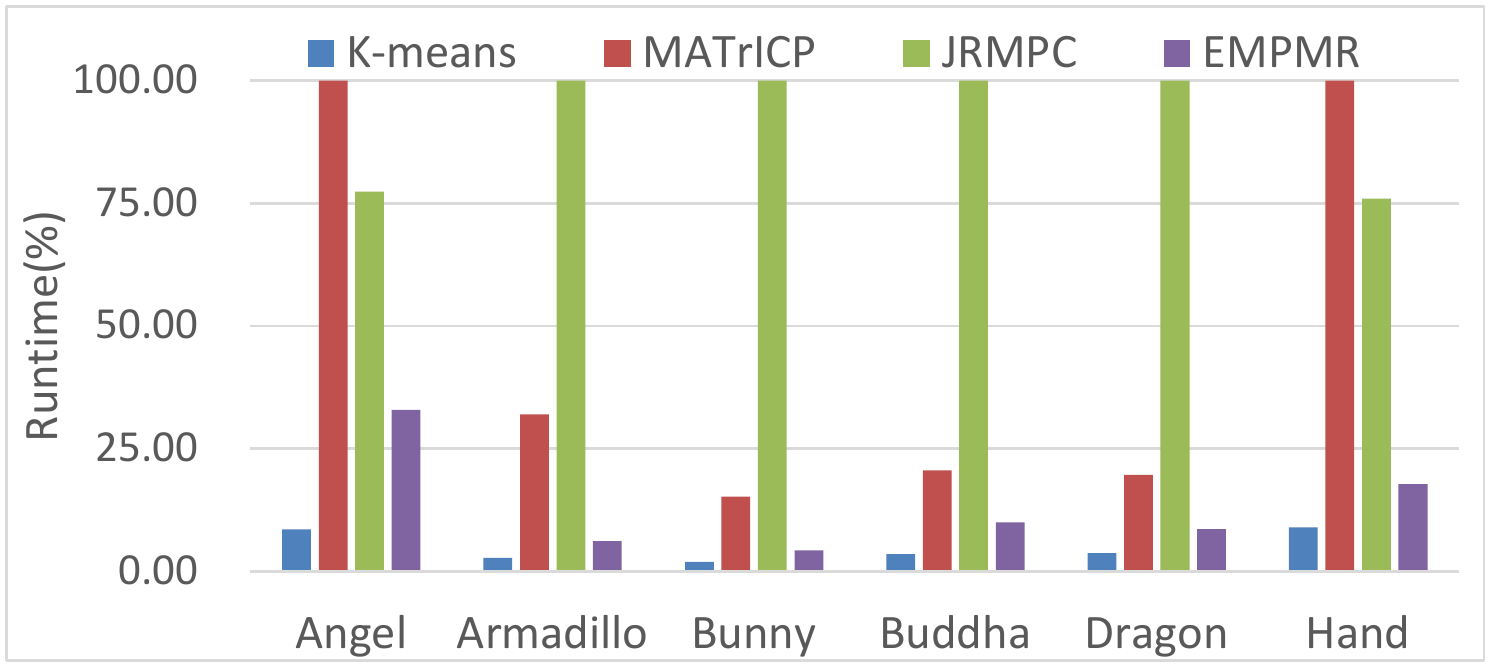}
}
\caption{ Comparison of average run time on six data sets. (a) Results under the noise of SNR=50dB, where [14.8358, 3.0061
2.4601, 3.669, 3.7294, 11.9983] minutes correspond to 100\% runtime of each data set. (b) Results under the noise of SNR=25dB, where [11.4680, 3.0056, 2.4868, 3.7176, 3.7298, 11.4713] minutes correspond to 100\% runtime of each data set.}
\label{Fig:MeanTime}
\end{figure*}

While JRMPC supposes that all points are drawn from a central Gaussian mixture and so cast the multi-view registration problem into a clustering problem. It utilizes the EM algorithm to simultaneously estimate all GMM parameters and rigid transformations that optimally align point sets. As a probabilistic method, this approach is expected to obtain accurate results in theoretically. However, it requires to estimate a huge number of parameters, whose initial values should be fine-tuned and provided in advance. Without good initial parameters, this approach is unable to obtain the desired registration results. Usually, it is difficult to provide good initial parameters for different data sets. Besides, the clustering in JRMPC also leads to the information loss, which can reduce the accuracy of registration results. Therefore, it is difficult to obtain promising results for multi-view registration in practice. As this approach requires to establish a large number of correspondences and estimate numerous components for the GMM, it is very time-consuming. These conclusions are also verified by experimental results illustrated in Table \ref{Tab:Acc}, Figs. \ref{Fig:Time} and \ref{Fig:Cros}.

\begin{figure*}[t]
\centering
\subfigure[]{
\includegraphics[scale=0.55]{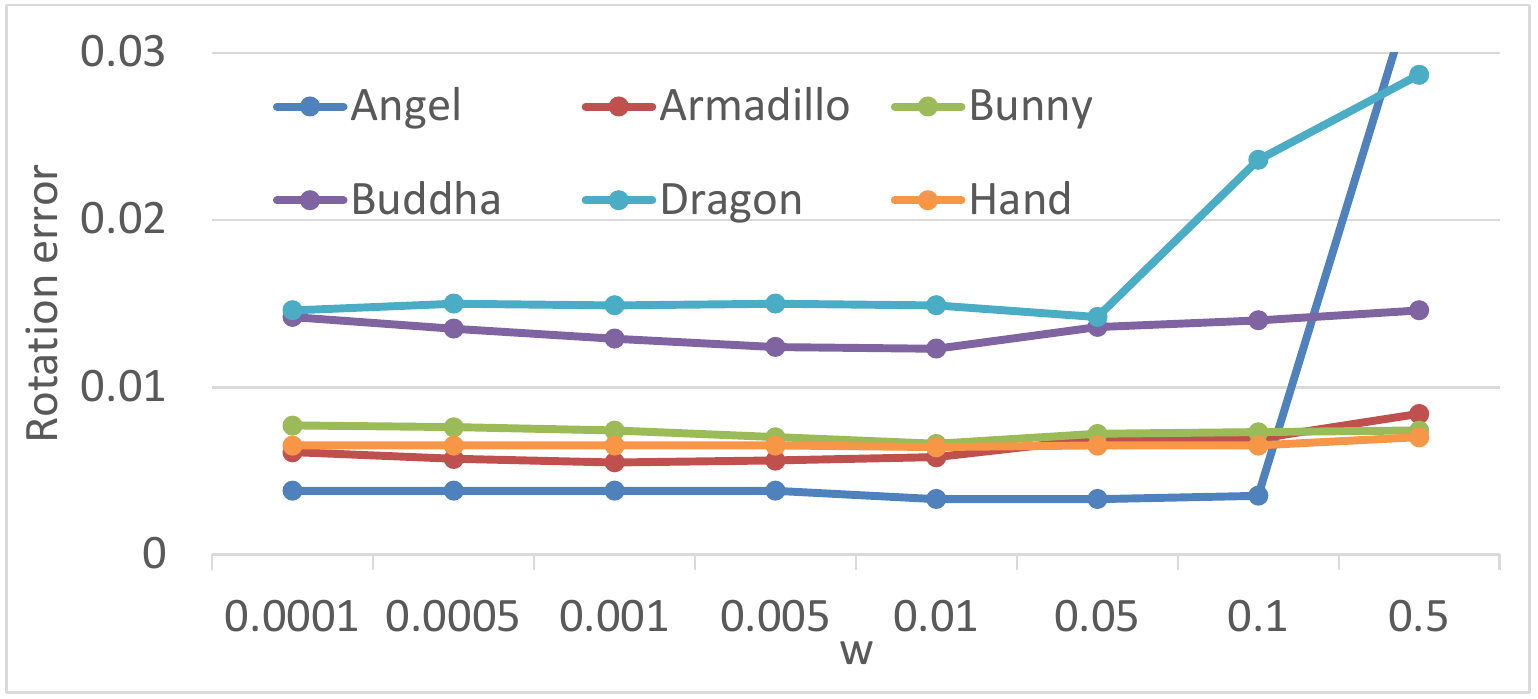}
}
\subfigure[]{
\includegraphics[scale=0.55]{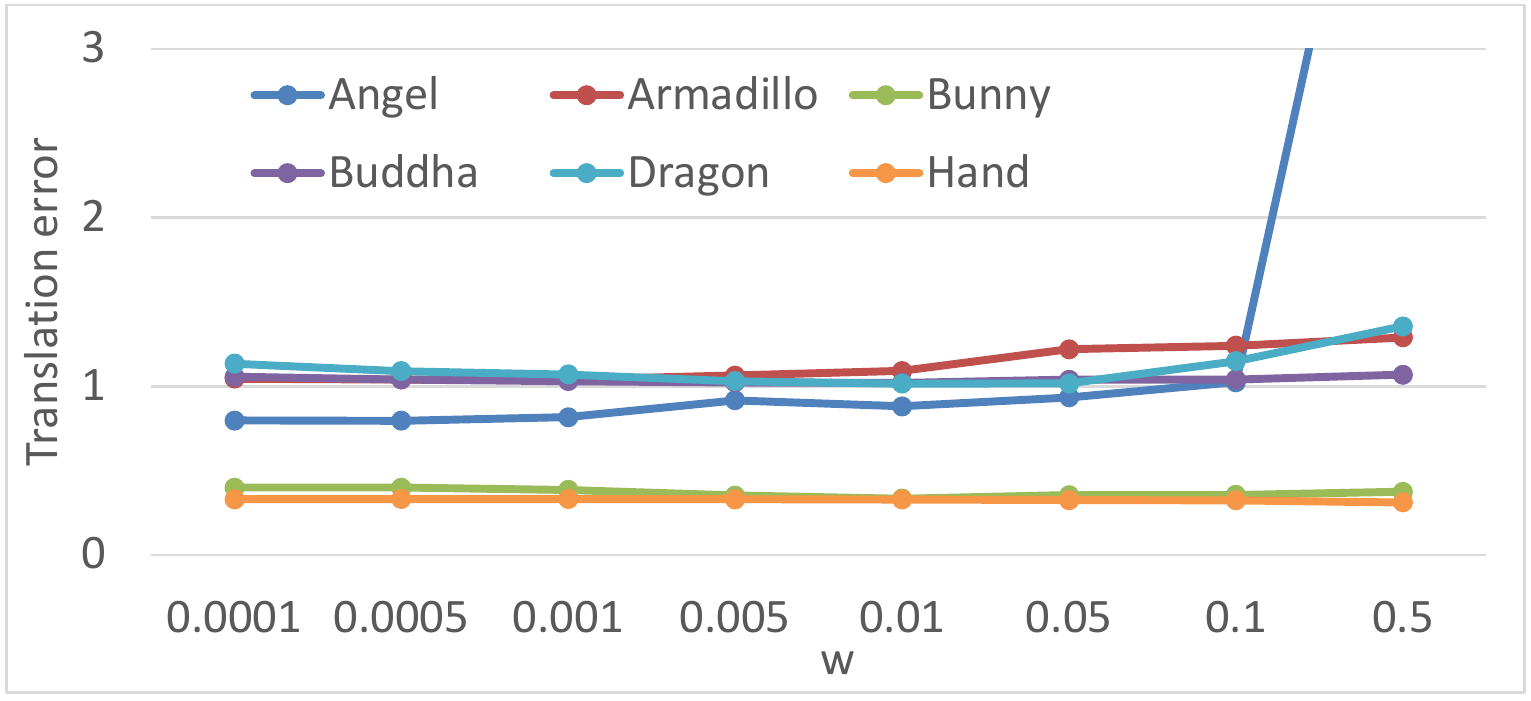}
}
\caption{Multi-view registration error of EMPMR under varied $w$. (a) Rotation errors. (b) Translation errors.}
\label{Fig:Para}
\end{figure*}

Different from JRMPC, EMPMR assumes that each data point is generated from one corresponding GMM, 
, whose centroids are specified by the NN search. As EMPMR utilizes equal covariances and equal membership probabilities for all GMM components, it only requires to estimate one covariance of as well as rigid transformations for multi-view registration. Besides, it only contains one preset parameter $w$, which can be set empirically. For each data point, this approach fully explores available information from all other opposite data sets by the NN search. Given appropriate initial parameters, it is more likely to obtain satisfactory results for the registration of multi-view point sets. Since all components of the GMM have been specified by the efficient NN search with one estimated covariance, this probabilistic method only requires to establish $(M-1)$ correspondences for each data point and estimate $M$ rigid transformations for the registration. Therefore, it is more efficient than JRMPC. In one word, EMPMR has good performance for the multi-view registration on both accuracy and efficiency.

\subsection {Robustness}
To illustrate its robustness, EMPMR is compared with other three competed approaches. They are tested on each data set, which is added with random Gaussian noises. To eliminate randomness, each group of experiment is carried out by 30 independent tests. Experimental results are also reported in the form of the average registration errors as well as the standard deviations. Tables \ref{Tab:50dB} and \ref{Tab:25dB} display all statistics results of all these competed methods under two levels of Gaussian noises. What's more, Fig. \ref{Fig:MeanTime} displays the averaging run time over 30 independent tests.

As shown in Tables \ref{Tab:50dB} and \ref{Tab:25dB}, the registration accruacy of all competed methods tends to be decreased with the increase of noise. Under the same level of noises, EMPMR is able to obtain the best registration results, except for the Buddha data set. For the Buddha data set, MATrICP can obtain the best performance and EMPMR is better than other two competed methods. As shown in Fig. \ref{Fig:MeanTime}, the efficiency of EMPMR is also comparable with CFTrICP, which is more efficient than other two competed approaches.

For most of noised data sets, the K-means based method can still obtain good registration results. But it is no longer able to maintain good registration for the Hand data sets. This is because the k-means clustering algorithm is sensitive to noises. Since this method utilizes the K-means clustering to achieve multi-view registration, its registration results are also affected by noises. Considering the stable registration results, MATrICP seems robust to noises for some data sets, i.e. Bunny, Buddha and Dragon data sets. As mentioned before, MATrICP recovers the multi-view registration results from a set of relative motions, which are estimated by the pair-wise registration algorithm.
Although the added noises may reduce the accuracy of pair-wise registration results, their influence on multi-view registration may be eliminated by the motion averaging algorithm. However, the added noise may result in the unreliable relative motions, which can lead to the undesired registration results for some data sets, such as Angel, Armadillo and Hand data sets. To achieve multi-view registration, both JRMPC and EMPMR utilizes the EM algorithm to estimate the GMM(s) as well as numerous rigid transformations. As the GMM(s) centriods get cleaned over time, these two methods seems robust to noises. However, JRMPC requires to estimate a huge number of GMM components as well as rigid transformations, which makes it tends to be trapped into local minimum. Therefore, JRMPC is different to obtain promising registration results. While, EMPMR utilizes the NN search to specify the centriods for all GMMs, it only requires to estimate $M$ rigid transformations as well as one GMM's covariance. As shown in Tables \ref{Tab:50dB}, \ref{Tab:25dB} and Fig \ref{Fig:MeanTime}, it is more likely to obtain promising registration results with good efficiency. Considering these facts, EMPMR is the most robust one among all these competed methods.

\subsection{Parameter sensitivity}
In EMPRM, there is a free parameter $w$, which requires to be set empirically. Thus, one question arising here is whether the performance of EMPRM is sensitive to $w$ or not. To answer this question, we conduct experiments on six data sets to observe the effects on multi-view registration performance with different values of $w$. Experimental results are reported in the form of registration errors, which are displayed in Fig. \ref{Fig:Para}.

Fig. \ref{Fig:Para} displays experimental results with different values of $\alpha$ on six data sets. It can be observed that: 1) the setting of $w \in [0.0005,0.05]$ is more likely to obtain accurate results for multi-view registration. 2) The performance of EMPRM only has small variations as long as $w$ is chosen in a suitable range, i.e., from 0.0005 to 0.05. Summarily, EMPRM is relatively insensitive to its parameter $w$ as long as it is chosen from a suitable range. This makes it easy to apply EMPRM without much effort for parameter tuning.

\begin{figure*}[!t]
\centering
\subfigure[]{\includegraphics[scale=1.0]{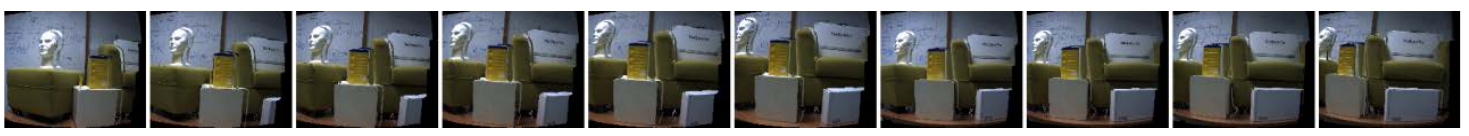}}
\vfill
\subfigure[]{\includegraphics[scale=0.46]{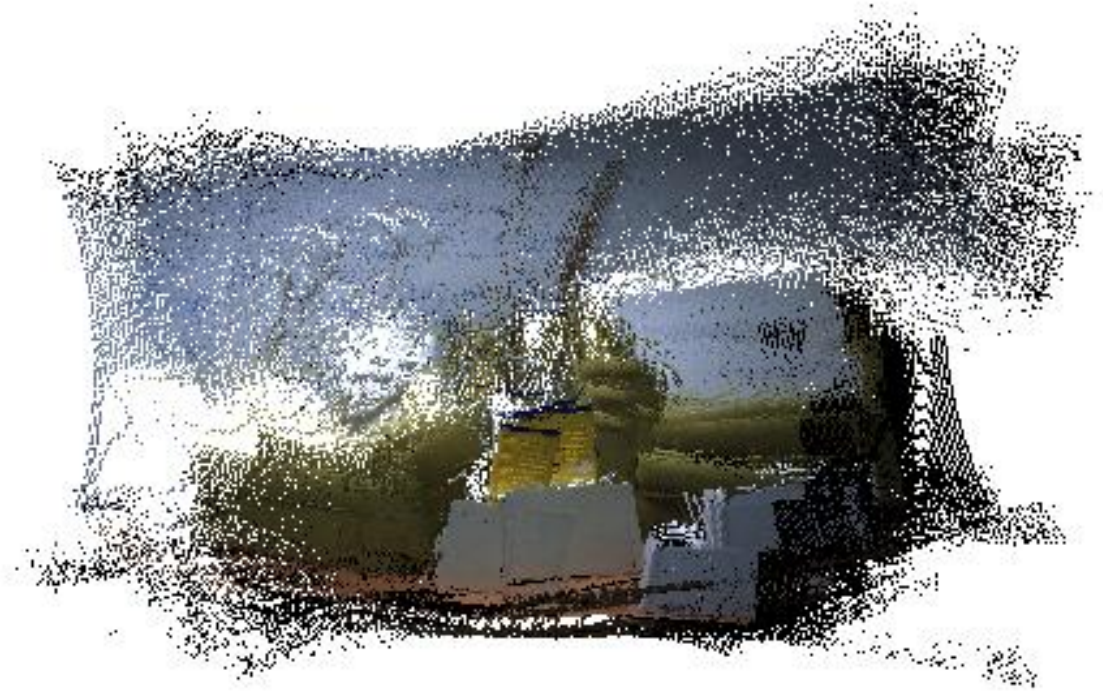}}
\subfigure[]{\includegraphics[scale=0.4]{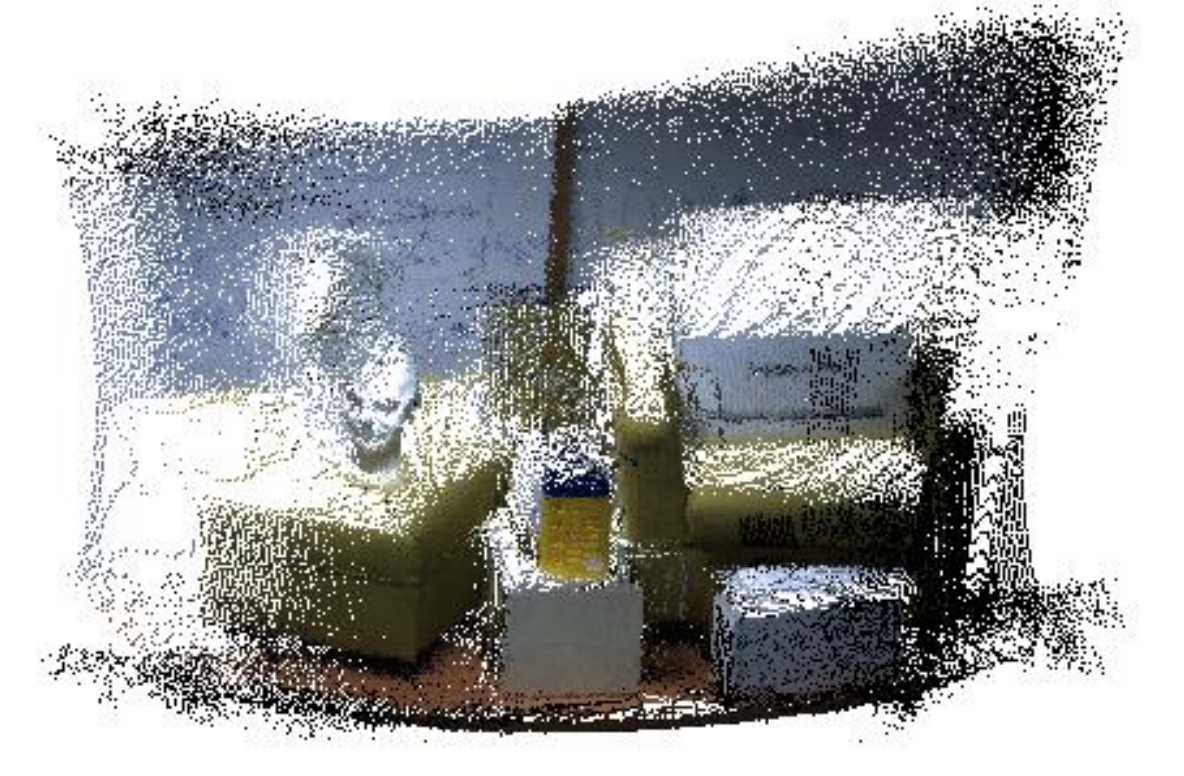}}
\subfigure[]{\includegraphics[scale=0.4]{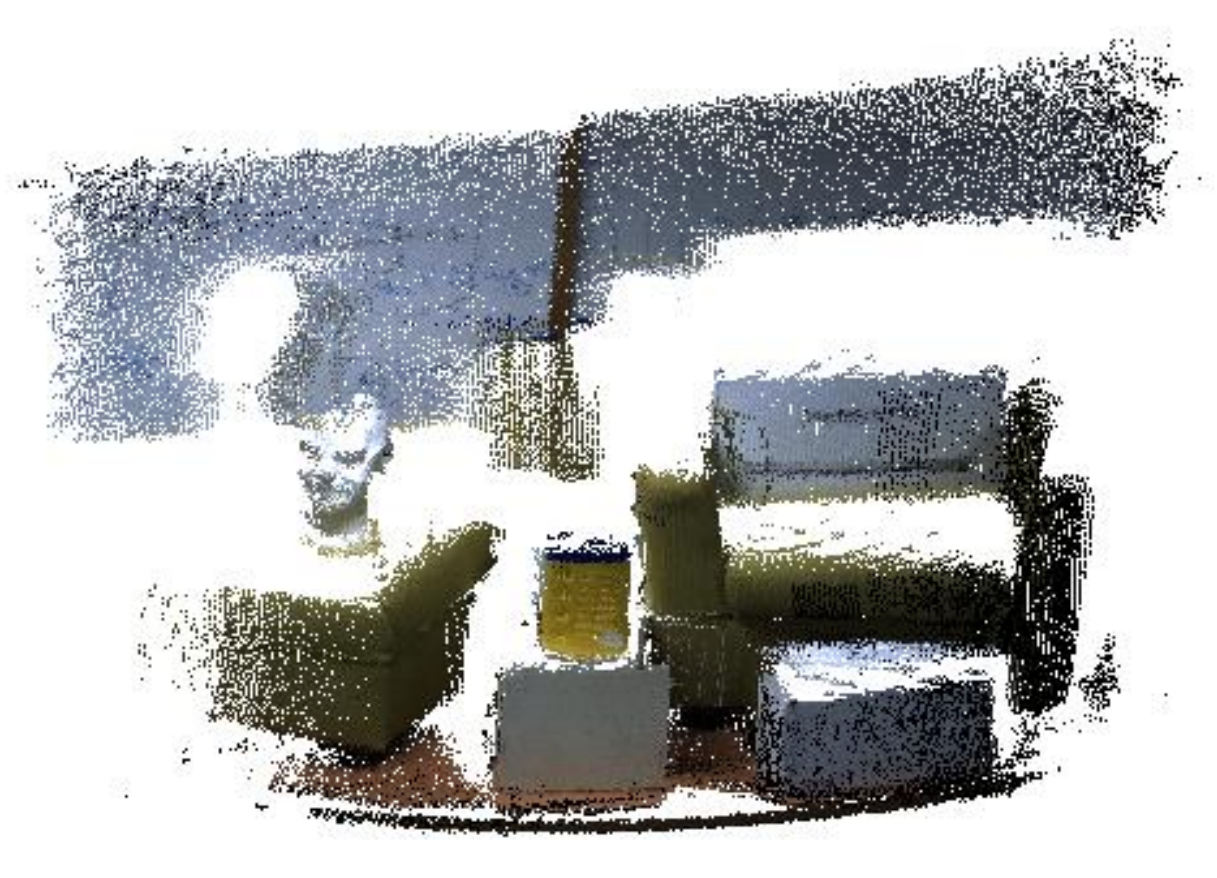}}
\caption{Scene reconstruction of the EXBI data set by EMPMR. (a) Scene image of each point set. (b) Initial aligned point sets. (c) Multi-view registration results with noise. (d) Scene reconstruction.}
\label{Fig:Scen}
\end{figure*}

\section{Scene reconstruction}
In practice, it is required to deal with different kinds of point sets. To illustrate its ability for the application, EMPMR is tested on the EXBI data set \cite{evangelidis2017joint} for the scene reconstruction.

The EXBI data set contains 10 RGB-D point sets recorded by the Kinect sensor, which acquires point clouds with associated color information. During the acquirement of this data set, the Kinect sensor was manually moving around an indoor scene. For the scene reconstruction, only information is processed by EMPMR and color information is utilized to assist for the final assessment. Given the initial rigid transformations, EMPMR is utilized to achieve accurate multi-view registration, which is the prerequisite for the scene reconstruction. Fig. \ref{Fig:Scen} displays the scene reconstruction of EMPMR. Since the raw RGB-D contains noises, the multi-view registration results should be filtered for good scene reconstruction. As shown in Fig. \ref{Fig:Scen}, EMPMR has the potential for the multi-view registration of RGB-D data and can be applied to the scene reconstruction.


\section{Conclusion}
Under the expectation-maximization perspective, this paper proposes an effective approach for the registration of multi-view point sets. In this method, 
each data point is assumed to be generated from one GMM, which is specified by all its NNs in other opposite point sets. In this way, the proposed method can treat all point sets on an equal footing: different points are a realization of the same number of GMMs and the multi-view registration is cast into a maximum likelihood estimation problem. To achieve the  multi-view registration, the EM-based algorithm is derived to maximize the likelihood function and estimate the rigid transformations for all point sets. Experimental results illustrate its superior performance over state-of-the-art approaches on accuracy robustness, and efficiency. What's more, this approach can be applied to 3D scene reconstruction. Our future work will extend this approach to solve the problem of robot 3D mapping.


\section*{Acknowledgment}
This work is supported by the National Natural Science Foundation of China under Grant nos. 61573273, in part by State Key Laboratory of Rail Transit Engineering Informatization (FSDI) under Grant Nos. SKLKZ19-01 and SKLK19-09. We also would like to thank Andrea Torsello for providing Angel and Hand datasets.

%
%
%

\bibliographystyle{IEEEtran}
\bibliography{ref}

\end{document}